# SVM and ELM: Who Wins? Object Recognition with Deep Convolutional Features from ImageNet

Lei Zhang and David Zhang

*Abstract*—Deep learning with a convolutional neural network (CNN) has been proved to be very effective in feature extraction and representation of images. For image classification problems, this work aim at finding which classifier is more competitive based on high-level deep features of images. In this report, we have discussed the nearest neighbor, support vector machines and extreme learning machines for image classification under deep convolutional activation feature representation. Specifically, we adopt the benchmark object recognition dataset from multiple sources with domain bias for evaluating different classifiers. The deep features of the object dataset are obtained by a well-trained CNN with five convolutional layers and three fully-connected layers on the challenging ImageNet. Experiments demonstrate that the ELMs outperform SVMs in cross-domain recognition tasks. In particular, state-of-the-art results are obtained by kernel ELM which outperforms SVMs with about 4% of the average accuracy. The features and codes are available in http://www.escience.cn/people/lei/index.html

## I. INTRODUCTION

Recently, deep learning as the hottest learning technique has been widely explored in machine learning, computer vision, natural language processing and data mining. In the early, convolutional neural network (CNN), as the most important deep net in deep learning, has been applied to document recognition and face recognition [1, 2]. Moreover, some deep learning algorithms with multi-layer fully connected networks (e.g. multi-layer perceptrons, MLP) for auto-encoder have been proposed, for examples, stacked auto encoders (SAE) [3], deep belief networks (DBN) [4] and deep Boltzmann machines (DBM) [5]. However, in large-scale learning problems, e.g. image classification in computer vision, CNNs with convolutioanl layers, pooling layers and fully-connected layers are widely investigated for its strong deep feature representation ability and state-of-the-art performance in challenged big datasets like ImageNet, Pascal VOC, etc. In the latest progress of deep learning, researchers have broken the new record in face verification by using CNNs with different structures [6, 7, 8, 9]. The latest verification accuracy on LFW data is 99.7% by Face++ team. Besides the faces, CNN has also achieved very competitive results on ImageNet for image classification and Pascal VOC data [10-17]. From these works, CNNs have been proved to be highly effective for deep feature representation with large-scale parameters. The main advantages of deep learning can be shown in three facets. 1) Feature representation. CNN integrates feature extraction (raw pixels) and model learning together, without using any other advanced low-level feature descriptors. 2) Large-scale learning. With the adjustable network structures, big data in millions can be learned by a CNN at one time. 3) Parameter learning. Due to the scalable network structures, millions of parameters can be trained. Therefore, CNN based deep method can be state-of-the-art parameter learning technique.

In this report, we would like to discuss about the deep feature representation capability of CNN by using traditional classification method with high-level deep features of images, and find which classifier is the best under the deep representation. Therefore, we mainly exploit the nearest neighbor (NN) [18], support vector machine (SVM) [19], least-square support vector machine (LSSVM) [20], extreme learning machine (ELM) [21] and kernel extreme learning machine (KELM) [22]. These classifiers are well-known in many different applications. Specially, ELM was initially proposed for generalized single-hidden-layer feed-forward neural networks and overcome the local minima, learning rate, stopping criteria and learning epochs that exist in gradient-based methods such as back-propagation (BP) algorithm. In recent years, ELMs are widely used due to some significant advantages such as learning speed, ease of implementation and minimal human intervention. The potential for large scale learning and artificial intelligence is preserved. The main steps of ELM include the random projection of hidden layer with random input weights and analytically determined solution by using Moore-Penrose generalized inverse. With similar impact with SVM, it has been proved to be efficient and effective for regression and classification tasks [23, 24]. The latest work about the principles and brain-alike learning of ELM has been presented [25]. Many improvement and new applications of ELMs have been proposed by researchers. The newest work about ELM for deep auto-encoder, local receptive fields for deep learning, transfer learning, and semi-supervised learning have also been proposed [26, 27, 28, 29, 30]. With the Mercer condition applied, a kernel ELM (KELM) that computes a kernel matrix of hidden layers has also been proposed [22]. A salient feature

L. Zhang is with College of Communication Engineering, Chongqing University, Chongqing 400044, China and also with Department of Computing, The Hong Kong Polytechnic University, Hong Kong (email: leizhang@cqu.edu.cn)
D. Zhang is with Department of Computing, The Hong Kong Polytechnic University, Hong Kong. (email: csdzhang@comp.polyu.edu.hk)

of KELM is that the random input weights and bias can be avoided.

In this report, we will present a study of NN, SVM, LSSVM, ELM and KELM for object recognition on the deep convolutional activation features trained by CNN on ImageNet, and have an insight of which one is the best for classification on deep representation.

The rest of this paper is organized as follows. Section 2 presents a method review of support vector machines and extreme learning machines. Section 3 shows the training and testing protocol of CNN for deep representation of images. Section 4 presents the experiments and results. Finally, Section 5 concludes this paper.

## II. OVERVIEW OF SVMs AND ELMs

### A. Support Vector Machine

In this section, the principle of SVM for classification problems is briefly reviewed. More details can be referred to [19].

Given a training set of $N$ data points $\{\mathbf{x}_i, y_i\}_{i=1}^{N}$, where the label $y_i \in \{-1, 1\}, i = 1, \cdots, N$. According to the structural risk minimization principle, SVM aims at solving the following risk bound minimization problem with inequality constraint.

$$\min_{\mathbf{w},\xi_i} \frac{1}{2}\|\mathbf{w}\|^2 + C \cdot \sum_{i=1}^{N} \xi_i, \quad (1)$$
$$s.t.\ \xi_i \geq 0,\ y_i\left[\mathbf{w}^T\varphi(\mathbf{x}_i)+b\right] \geq 1-\xi_i$$

where $\varphi(\cdot)$ is a linear/nonlinear mapping function, $\mathbf{w}$ and $b$ are the parameters of classifier hyper-plane.

Generally, for optimization, the original problem (7) of SVM can be transformed into its dual formulation with equality constraint by using Lagrange multiplier method. One can construct the Lagrange function

$$L(\mathbf{w},b,\xi_i;\alpha_i,\lambda_i) = \frac{1}{2}\|\mathbf{w}\|^2 + C \cdot \sum_{i=1}^{N} \xi_i \\ - \sum_{i=1}^{N} \alpha_i\left(y_i\left[\mathbf{w}^T\varphi(\mathbf{x}_i)+b\right]-1+\xi_i\right) - \sum_{i=1}^{N}\lambda_i\xi_i \quad (2)$$

where $\alpha_i \geq 0$ and $\lambda_i \geq 0$ are Lagrange multipliers. The solution can be given by the saddle point of Lagrange function (2) by solving

$$\max_{\alpha_i,\lambda_i} \min_{\mathbf{w},b,\xi_i} L(\mathbf{w},b,\xi_i;\alpha_i,\lambda_i) \quad (3)$$

By calculating the partial derivatives of Lagrange function (2) with respect to $\mathbf{w}$, $b$ and $\xi_i$, one can obtain

$$\begin{cases} \dfrac{\partial L(\mathbf{w},b,\xi_i;\alpha_i,\lambda_i)}{\partial \mathbf{w}} = 0 \rightarrow \mathbf{w} = \sum_{i=1}^{N}\alpha_i y_i \varphi(\mathbf{x}_i) \\ \dfrac{\partial L(\mathbf{w},b,\xi_i;\alpha_i,\lambda_i)}{\partial b} = 0 \rightarrow \sum_{i=1}^{N}\alpha_i y_i = 0 \\ \dfrac{\partial L(\mathbf{w},b,\xi_i;\alpha_i,\lambda_i)}{\partial \xi_i} = 0 \rightarrow 0 \leq \alpha_i \leq C \end{cases} \quad (4)$$

Then one can rewrite (3) as

$$\max_{\boldsymbol{\alpha}} \sum_i \boldsymbol{\alpha}_i - \frac{1}{2}\sum_{i,j} y_i y_j \boldsymbol{\alpha}_i \boldsymbol{\alpha}_j \varphi(\mathbf{x}_i)^T \varphi(\mathbf{x}_j) \quad (5)$$
$$s.t.\ \sum_{i=1}^{N}\alpha_i y_i = 0,\ 0 \leq \alpha_i \leq C$$

By solving α of the dual problem (5) with a quadratic programming, the goal of SVM is to construct the following decision function (classifier),

$$f(\mathbf{x}) = \text{sgn}\left(\sum_{i=1}^{M}\boldsymbol{\alpha}_i y_i \kappa(\mathbf{x}_i, \mathbf{x}) + b\right) \quad (6)$$

where $\kappa(\cdot)$ is a kernel function. $\kappa(\mathbf{x}_i, \mathbf{x}) = \varphi(\mathbf{x}_i)^T\varphi(\mathbf{x}) = \mathbf{x}_i^T\mathbf{x}$ for linear SVM and $\kappa(\mathbf{x}_i, \mathbf{x}) = \exp(-\|\mathbf{x}_i - \mathbf{x}\|^2/\sigma^2)$ for RBF-SVM.

### B. Least Square Support Vector Machine

LSSVM is an improved and simplified version of SVM. The details can be referred to [20]. We briefly introduce the basic principle of LSSVM for classification problems. By introducing the square error and equality constraint, LSSVM can be formulated as

$$\min_{\mathbf{w},\xi_i} \frac{1}{2}\|\mathbf{w}\|^2 + C \cdot \frac{1}{2}\sum_{i=1}^{N}\xi_i^2, \quad (7)$$
$$s.t.\ y_i\left[\mathbf{w}^T\varphi(\mathbf{x}_i)+b\right] = 1-\xi_i,\ i=1,\ldots,N$$

The Lagrange function of (7) can be defined as

$$L(\mathbf{w},b,\xi_i;\alpha_i) = \frac{1}{2}\|\mathbf{w}\|^2 + C \cdot \frac{1}{2}\sum_{i=1}^{N}\xi_i^2 \\ - \sum_{i=1}^{N}\alpha_i\left(y_i\left[\mathbf{w}^T\varphi(\mathbf{x}_i)+b\right]-1+\xi_i\right) \quad (8)$$

where $\alpha_i$ is the Lagrange multiplier.

The optimality conditions can be obtained by computing the partial derivatives of (8) with respect to the four variables as

$$\begin{cases} \dfrac{\partial L(\mathbf{w},b,\xi_i;\alpha_i)}{\partial \mathbf{w}} = 0 \rightarrow \mathbf{w} = \sum_{i=1}^{N}\alpha_i y_i \varphi(\mathbf{x}_i) \\ \dfrac{\partial L(\mathbf{w},b,\xi_i;\alpha_i)}{\partial b} = 0 \rightarrow \sum_{i=1}^{N}\alpha_i y_i = 0 \\ \dfrac{\partial L(\mathbf{w},b,\xi_i;\alpha_i)}{\partial \xi_i} = 0 \rightarrow \alpha_i = C\xi_i \\ \dfrac{\partial L(\mathbf{w},b,\xi_i;\alpha_i)}{\partial \alpha_i} = 0 \rightarrow y_i\left[\mathbf{w}^T\varphi(\mathbf{x}_i)+b\right]-1+\xi_i = 0 \end{cases} \quad (9)$$

The equation group (9) can be written in linear equation as

$$\begin{bmatrix} \mathbf{I} & 0 & 0 & -\mathbf{Z}^T \\ 0 & 0 & 0 & -\mathbf{Y}^T \\ 0 & 0 & C\mathbf{I} & -\mathbf{I} \\ \mathbf{Z} & \mathbf{Y} & \mathbf{I} & 0 \end{bmatrix} \begin{bmatrix} \mathbf{w} \\ b \\ \boldsymbol{\xi} \\ \boldsymbol{\alpha} \end{bmatrix} = \begin{bmatrix} \mathbf{0} \\ 0 \\ \mathbf{0} \\ \vec{\mathbf{1}} \end{bmatrix} \quad (10)$$

where $\mathbf{Z} = [\varphi(\mathbf{x}_1)y_1,\ldots,\varphi(\mathbf{x}_N)y_N]^T$, $\mathbf{Y} = [y_1,\ldots,y_N]^T$, $\vec{\mathbf{1}} = [1,\ldots,1]^T$, $\boldsymbol{\xi} = [\xi_1,\ldots,\xi_N]^T$, $\boldsymbol{\alpha} = [\alpha_1,\ldots,\alpha_N]^T$. The solution

of **α** and **b** can also be given by

$$\begin{bmatrix} 0 & -\mathbf{Y}^T \\ \mathbf{Y} & \mathbf{ZZ}^T + C^{-1}\mathbf{I} \end{bmatrix} \begin{bmatrix} \mathbf{b} \\ \mathbf{\alpha} \end{bmatrix} = \begin{bmatrix} 0 \\ \vec{1} \end{bmatrix} \quad (11)$$

Let $\mathbf{\Omega} = \mathbf{ZZ}^T$, with the Mercer condition, there is

$$\Omega_{k,l} = y_k y_l \varphi(\mathbf{x}_k)^T \varphi(\mathbf{x}_l) = y_k y_l \kappa(\mathbf{x}_k, \mathbf{x}_l), k, l = 1, \ldots, N \quad (12)$$

By substituting (12) into (11), the solution can be obtained by solving a linear equation instead of a quadratic programming problem in SVM. The final decision function of LSSVM is the same as SVM shown as (6).

*C. Extreme Learning Machine*

ELM aims to solve the output weights of a single layer feed-forward neural network (SLFN) by minimizing the squared loss of predicted errors and the norm of the output weights in both classification and regression problems. We briefly introduce the principle of ELM for classification problems. Given a dataset $\mathbf{X} = [\mathbf{x}_1, \mathbf{x}_2, \cdots, \mathbf{x}_N] \in \Re^{d \times N}$ of $N$ samples with label $\mathbf{T} = [\mathbf{t}_1, \mathbf{t}_2, \cdots, \mathbf{t}_N] \in \Re^{c \times N}$, where $d$ is the dimension of sample and $c$ is the number of classes. Note that if $\mathbf{x}_i$ $(i = 1, \cdots, N)$ belongs to the $k$-th class, the $k$-th position of $\mathbf{t}_i$ $(i = 1, \cdots, N)$ is set as 1, and -1 otherwise. The hidden layer output matrix $\mathbf{H}$ with $L$ hidden neurons can be computed as

$$\mathbf{H} = \begin{bmatrix} h(\mathbf{w}_1^T \mathbf{x}_1 + b_1) & h(\mathbf{w}_2^T \mathbf{x}_1 + b_2) & \cdots & h(\mathbf{w}_L^T \mathbf{x}_1 + b_L) \\ \vdots & \vdots & & \vdots \\ h(\mathbf{w}_1^T \mathbf{x}_N + b_1) & h(\mathbf{w}_2^T \mathbf{x}_N + b_2) & \cdots & h(\mathbf{w}_L^T \mathbf{x}_N + b_L) \end{bmatrix} \quad (13)$$

where $h(\cdot)$ is the activation function of hidden layer, $\mathbf{W} = [\mathbf{w}_1, \cdots, \mathbf{w}_L] \in \Re^{d \times L}$ and $\mathbf{B} = [b_1, \cdots, b_L]^T \in \Re^L$ are randomly generated input weights and bias between the input layer and hidden layer. With such a hidden layer output matrix $\mathbf{H}$, ELM can be formulated as follows

$$\min_{\mathbf{\beta} \in \Re^{L \times c}} \frac{1}{2} \|\mathbf{\beta}\|^2 + C \cdot \frac{1}{2} \sum_{i=1}^N \|\mathbf{\xi}_i\|^2$$
$$s.t. \; h(\mathbf{x}_i)\mathbf{\beta} = \mathbf{t}_i^T - \mathbf{\xi}_i^T, i = 1, \ldots, N \Leftrightarrow \mathbf{H}\mathbf{\beta} = \mathbf{T}^T - \mathbf{\xi}^T \quad (14)$$

where $\mathbf{\beta} \in \Re^{L \times c}$ denotes the output weights between hidden layer and output layer, $\mathbf{\xi} = [\mathbf{\xi}_1, \cdots, \mathbf{\xi}_N]$ denotes the prediction error matrix with respect to the training data, and $C$ is a penalty constant on the training errors.

The closed form solution $\mathbf{\beta}$ of (14) can be easily solved. First, if the number $N$ of training patterns is larger than $L$, the gradient equation is over-determined, and the closed form solution of (14) can be obtained as

$$\mathbf{\beta}^* = \mathbf{H}^+ \mathbf{T} = \left( \mathbf{H}^T \mathbf{H} + \frac{\mathbf{I}_{L \times L}}{C} \right)^{-1} \mathbf{H}^T \mathbf{T} \quad (15)$$

where $\mathbf{I}_{L \times L}$ denotes the identity matrix with size of $L$, and $\mathbf{H}^+$ is the Moore-Penrose generalized inverse of $\mathbf{H}$.

If the number $N$ of training patterns is smaller than $L$, an under-determined least square problem would be handled. In this case, the solution of (14) can be obtained as

$$\mathbf{\beta}^* = \mathbf{H}^+ \mathbf{T} = \mathbf{H}^T \left( \mathbf{H}\mathbf{H}^T + \frac{\mathbf{I}_{N \times N}}{C} \right)^{-1} \mathbf{T} \quad (16)$$

where $\mathbf{I}_{N \times N}$ denotes the identity matrix.
Then the predicted output of a new observation $\mathbf{z}$ can be computed as

$$\mathbf{y} = h(\mathbf{z})\mathbf{\beta}^* = \begin{cases} h(\mathbf{z}) \cdot \left( \mathbf{H}^T \mathbf{H} + \frac{\mathbf{I}_{L \times L}}{C} \right)^{-1} \mathbf{H}^T \mathbf{T}, & \text{if } N \geq L \\ h(\mathbf{z}) \cdot \mathbf{H}^T \left( \mathbf{H}\mathbf{H}^T + \frac{\mathbf{I}_{N \times N}}{C} \right)^{-1} \mathbf{T}, & \text{if } N < L \end{cases} \quad (17)$$

*D. Kernelized Extreme Learning Machine*

One can also apply Mercer condition to ELM and thus a KELM is formulated. The KELM can be described as follows. Let $\mathbf{\Omega} = \mathbf{H}\mathbf{H}^T \in \Re^{N \times N}$, where $\Omega_{i,j} = h(\mathbf{x}_i)h(\mathbf{x}_j)^T = \kappa(\mathbf{x}_i, \mathbf{x}_j)$ and $\kappa(\cdot)$ is the kernel function. With the expression of solution $\mathbf{\beta}$ (16), the predicted output of a new observation $\mathbf{z}$ can be computed as

$$\begin{aligned} \mathbf{y} &= h(\mathbf{z})\mathbf{\beta}^* \\ &= h(\mathbf{z}) \cdot \mathbf{H}^T \left( \mathbf{H}\mathbf{H}^T + \frac{\mathbf{I}_{N \times N}}{C} \right)^{-1} \mathbf{T} \\ &= \begin{bmatrix} \kappa(\mathbf{z}, \mathbf{x}_1) \\ \vdots \\ \kappa(\mathbf{z}, \mathbf{x}_1) \end{bmatrix}^T \left( \mathbf{\Omega} + \frac{\mathbf{I}_{N \times N}}{C} \right)^{-1} \mathbf{T} \end{aligned} \quad (18)$$

Note that due to the kernel matrix of training data is $\mathbf{\Omega} \in \Re^{N \times N}$, therefore, the number $L$ of hidden neurons is not explicit and the decision function of KELM can be expressed uniquely in (18).

III. TRAINING AND TESTING PROTOCOL

*A. CNN training on ImageNet*

In this report, we aim at proposing a comparative investigation on SVMs and ELMs for classification based on deep convolutional features. Therefore, we adopt the deep convolutional activated features (DeCAF) from [17] for experiments. The structures of CNN for training on the ImageNet with 1000 categories are the same as the proposed CNN in [10]. The basic structure of the adopted is illustrated in Fig.1, which includes 5 convolutional layers and 3 fully-connected layers. Further details of the CNN training architecture and features can be referred to [10, 17].

TABLE I
DETAILS OF 4DA-CNN DATASETS

| Dataset | #class | #dimension | #samples | $n_s/c$ | $n_t/c$ |
|---|---|---|---|---|---|
| Amazon | 10 | 4096 | 958 | 20 | 3 |
| DSLR | 10 | 4096 | 157 | 8 | 3 |
| Webcam | 10 | 4096 | 295 | 8 | 3 |
| Caltech | 10 | 4096 | 1123 | 8 | 3 |

## B. CNN Testing

The well-trained network parameters shown in Fig.1 are used for deep representation of the 4DA (domain adaptation) dataset [31, 32]. The CNN outputs of the 6-th ($f_6$) and 7-th ($f_7$) fully-connected layers are used as inputs of SVMs and ELMs for classification, respectively. The 4DA dataset includes four domains such as Caltech 256 (C), Amazon (A), Webcam (W) and Dslr (D) sampled from different sources, in which 10 object classes are selected. As can be seen from Fig.1, the dimension of features from $f_6$ and $f_7$ is 4096. The detail of 4DA dataset with deep features is summarized in Table I. Some examples of the dataset for each domain have been illustrated in Fig.2.

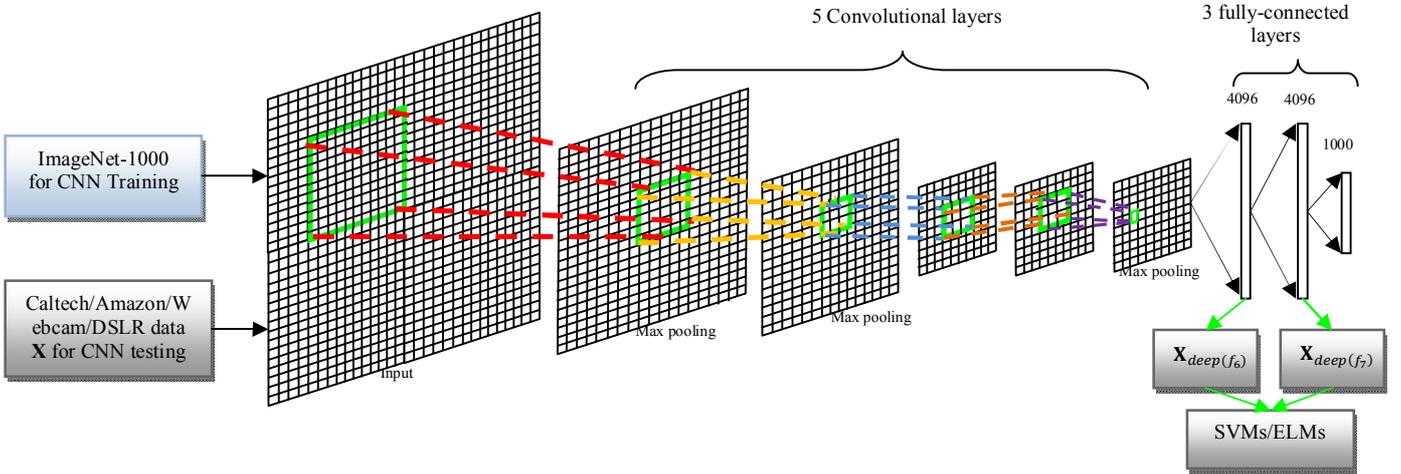

Fig. 1. Diagram of the training and testing protocol in this report

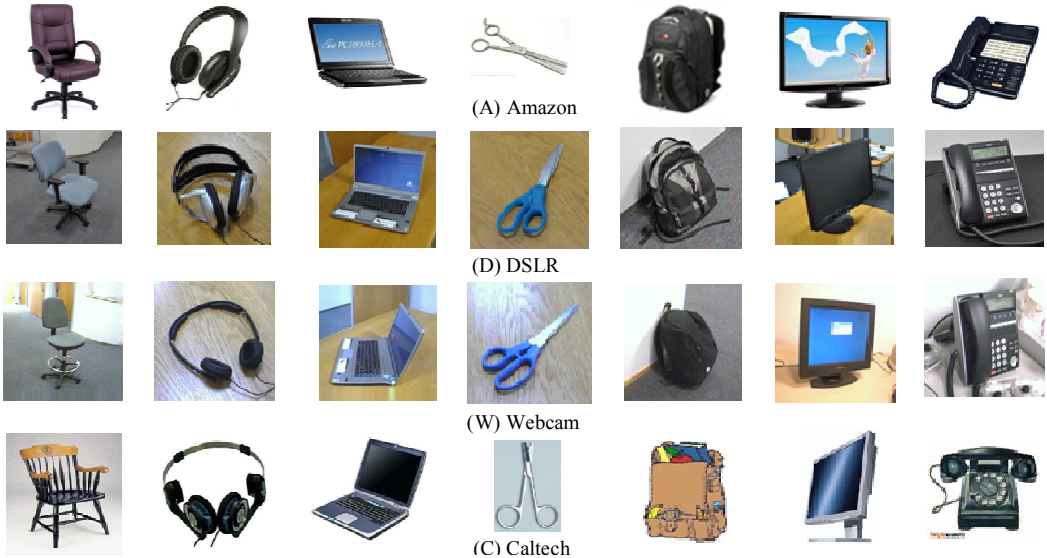

Fig. 2. Examples of object images from three sources: Amazon (1st row), DSLR (2nd row), Webcam (3rd row) and Caltech 256 (4th row). Different visual cues such as camera viewpoint, resolution, illumination, and background have been well illustrated.

## C. Classification

The 4DA dataset is commonly used for evaluating domain adaptation and transfer learning tasks. So, in this report, we investigate the classification ability of deep representation on domain shifted data. We adopt the deep features for SVMs/ELMs training, and compare the classification accuracy. The specific experimental setup is described in Experiments section.

## IV. EXPERIMENTS

### A. Experimental Setup

In the experiment, three settings are investigated respectively, as follows.

1) **Setting 1:** *single-domain* recognition task.

For example, we train a model on the training data of Amazon, and report the test accuracy on the remaining data of Amazon. As shown in Table I ($n_s/c$), 20, 8, 8, and 8 samples per class are randomly selected for training from Amazon, DSLR, Webcam and Caltech domains, respectively, and the remaining are used as test samples for each domain. 20 random train/test splits are run, and the average recognition accuracy for each method is reported.

2) **Setting 2:** *cross-domain* recognition tasks--source only.

We perform a cross-domain recognition task. For example, we train a SVM/ELM on the Amazon and test on DSLR, i.e.

A→D. Totally, 12 cross-domain tasks among the four domains are conducted. Note that the training data is source data only (source only) without leveraging the data from target domain. The number of training data is 20, 8, 8 and 8 per class for Amazon, DSLR, Webcam and Caltech domains, respectively, when used as source domain. 20 random train/test splits are run, and the average recognition accuracy for each method is reported.

3) **Setting 3:** *cross-domain* recognition tasks--source and target.

Similar to Setting 2, we perform a cross-domain recognition task. For example, we train a SVM/ELM on the Amazon and test on DSLR, i.e. A→D. Totally, 12 cross-domain tasks among the four domains are conducted. However, the difference from Setting 2 lies in that the training data includes the labeled source data and few labeled target data. The number of training data is 20, 8, 8 and 8 per class for Amazon, DSLR, Webcam and Caltech domains, respectively, when used as source domain. The number of few labeled target data is 3 per class for each domain when they are used as target domain, as shown in Table I ($n_t/c$). 20 random train/test splits are run, and the average recognition accuracy for each method is reported.

### B. Parameter Setting

To make sure that the best result of each method can be obtained, we have adjusted the parameters. For SVM the penalty coefficient $C$ and kernel parameter $\sigma$ are set as 1000 and 1, respectively, by using *Libsvm-3.12* toolbox. For LSSVM, the two coefficients are automatically optimized with a grid search by using *LSSVM-1.7* toolbox. For ELM, the penalty coefficient $C$ and the number $L$ of hidden neurons are set as 100 and 5000, respectively. For KELM, the penalty coefficient $C$ and kernel parameter $\sigma$ are set as 100 and 0.01, respectively. Note that the penalty coefficient C and kernel parameter $\sigma$ for SVM, ELM, and KELM are adjusted from the set $C=\{1, 100, 10000\}$ and $\sigma=\{0.0001, 0.01, 1, 100\}$.

### C. Experimental Results

(1) Results of **Setting 1**.

For experimental **Setting 1**, the average accuracy of 20 randomly generated train/test splits for five methods including NN, SVM, LSSVM, ELM and KELM are reported in Table II. We can observe that the recognition performance based on the deep features from the 6-th layer ($f_6$) and 7-th layer ($f_7$) is slightly different. The best two methods are highlighted with bold face. From the comparisons, we can find that ELMs outperforms SVMs and NN methods for all domains, and KELM shows a more competitive performance. Specifically, by comparing KELM and SVM, the improvement in accuracy for the deep features $f_6$ is 0.8%, 0.2%, 1.1% and 2.1% for Amazon, DSLR, Webcam, and Caltech, respectively. For the deep features $f_7$, the improvement is 1.0%, 0.6%, 0.8%, and 2.5%, respectively.

TABLE II
RECOGNITION ACCURACY OF EACH METHOD FOR DIFFERENT DOMAINS IN **SETTING 1**

| Method | CNN_layer | Amazon | DSLR | Webcam | Caltech | CNN_layer | Amazon | DSLR | Webcam | Caltech |
|---|---|---|---|---|---|---|---|---|---|---|
| NN | $f_6$ | 91.0±0.3 | 97.3±0.6 | 95.0±0.4 | 75.0±0.4 | $f_7$ | 92.4±0.2 | 96.8±0.5 | 95.3±0.5 | 76.2±0.5 |
| SVM | $f_6$ | **92.9±0.1** | 97.6±0.6 | 96.7±0.3 | 83.9±0.4 | $f_7$ | 93.2±0.1 | 96.9±0.5 | 96.5±0.4 | 83.2±0.5 |
| LSSVM | $f_6$ | **92.9±0.2** | 97.5±0.4 | 96.4±0.4 | 84.6±0.3 | $f_7$ | 93.5±0.1 | 96.3±0.6 | 95.4±0.4 | 83.9±0.4 |
| ELM | $f_6$ | 92.9±0.1 | **98.0±0.3** | **97.7±0.2** | **84.8±0.3** | $f_7$ | **93.6±0.1** | **97.2±0.4** | **97.4±0.3** | **85.0±0.3** |
| KELM | $f_6$ | **93.7±0.1** | **97.8±0.3** | **97.8±0.2** | **86.0±0.3** | $f_7$ | **94.2±0.1** | **97.5±0.4** | **97.3±0.4** | **85.7±0.3** |

(2) Results of **Setting 2**.

Table III presents the average recognition accuracy of 20 randomly generated train/test splits based on the experimental setting 2. Totally, 12 cross-domain recognition tasks are conducted. The first two highest accuracies are highlighted in bold face. We can observe that 1) the recognition performance with deep feature $f_7$ clearly outperforms that of $f_6$, which demonstrates the effectiveness of "deep"; 2) the performance of ELM and KELM is significantly better than SVM and LSSVM, the average improvement of 12 tasks of KELM is 4% better than that of SVM. The results demonstrate that for more difficult problems (i.e. cross-domain tasks), the ELM based methods show a more competitive and robust advantage for classification. More obvious, the accuracies by using the five methods for each cross-domain task are illustrated in Fig. 3, from which the superiority of ELMs especially KELM is clearly demonstrated compared with others methods for each tasks under deep features from $f_7$ and $f_6$.

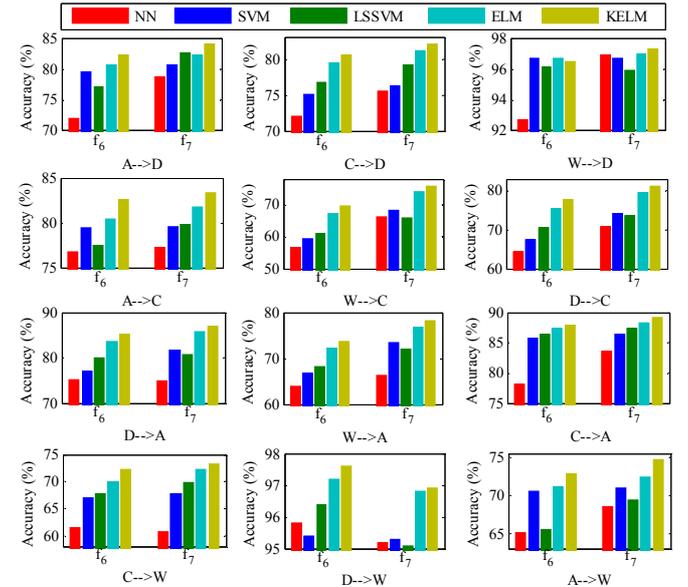

Fig. 3. Recognition accuracies of 12 cross-domain tasks by using NN, SVM, LSSVM, ELM and KELM on the deep convolutional activation features of $f_6$ and $f_7$ with experimental **Setting 2**

TABLE III
RECOGNITION ACCURACY OF EACH METHOD WITH **SETTING 2**, WHERE THE TRAINING DATA IS FROM SOURCE DOMAIN ONLY (A: AMAZON, C: CALTECH 256, W: WEBCAM, D: DSLR)

| Method | CNN_layer | A→D | C→D | W→D | A→C | W→C | D→C | D→A | W→A | C→A | C→W | D→W | A→W |
|---|---|---|---|---|---|---|---|---|---|---|---|---|---|
| NN | $f_6$ | 71.9±0.9 | 72.0±1.7 | 92.7±0.5 | 76.8±0.3 | 56.6±0.9 | 64.4±0.4 | 75.1±0.7 | 64.0±0.6 | 78.1±0.8 | 61.5±1.1 | 95.8±0.4 | 65.1±1.0 |
| | $f_7$ | 78.7±0.5 | 75.6±1.3 | 96.9±0.4 | 77.2±0.4 | 66.2±0.5 | 70.7±0.4 | 75.0±0.7 | 66.3±0.8 | 83.6±0.4 | 60.7±1.2 | 95.2±0.4 | 68.5±0.8 |
| SVM | $f_6$ | 79.6±0.7 | 75.1±1.8 | 96.7±0.4 | 79.5±0.4 | 59.5±0.9 | 67.3±1.2 | 77.0±1.0 | 66.8±1.0 | 85.8±0.4 | 67.1±1.1 | 95.4±0.4 | 70.6±0.8 |
| | $f_7$ | 80.6±0.8 | 76.4±1.4 | 96.7±0.4 | 79.6±0.4 | 68.1±0.6 | 74.3±0.6 | 81.8±0.5 | 73.4±0.7 | 86.5±0.5 | 67.8±1.1 | 95.3±0.5 | 71.0±0.8 |
| LSSVM | $f_6$ | 77.1±0.9 | 76.8±1.2 | 96.1±0.3 | 77.5±0.6 | 61.1±0.7 | 70.6±1.0 | 80.0±0.8 | 68.2±1.1 | 86.5±0.4 | 67.8±1.2 | 96.4±0.4 | 65.5±0.8 |
| | $f_7$ | **82.6±0.5** | 79.2±0.8 | 95.9±0.4 | 79.8±0.5 | 66.0±1.3 | 73.7±0.9 | 80.8±0.7 | 72.0±1.1 | 87.4±0.3 | 69.9±1.1 | 95.1±0.3 | 69.4±0.6 |
| ELM | $f_6$ | 80.6±0.6 | 79.5±1.2 | 96.7±0.2 | 80.4±0.3 | 67.2±0.5 | 75.6±0.5 | 83.7±0.4 | 72.2±0.9 | 87.3±0.4 | 70.1±0.9 | **97.2±0.3** | 71.1±0.6 |
| | $f_7$ | 82.3±0.5 | **81.2±0.7** | **97.0±0.4** | 81.8±0.3 | **74.0±0.3** | 79.5±0.2 | 85.8±0.3 | 76.7±0.2 | 88.3±0.2 | 72.3±0.9 | 97.0±0.3 | 72.4±0.8 |
| KELM | $f_6$ | 82.3±0.5 | 80.7±0.9 | 96.5±0.3 | **82.6±0.3** | 69.5±0.4 | 77.8±0.4 | 85.3±0.4 | 73.8±1.1 | 88.0±0.4 | 72.3±1.0 | 97.6±0.2 | **72.9±0.7** |
| | $f_7$ | **84.0±0.4** | **82.2±0.9** | **97.3±0.3** | **83.4±0.2** | **75.7±0.3** | **81.1±0.2** | **87.1±0.2** | **78.2±0.8** | **89.1±0.3** | **73.3±0.9** | 96.9±0.3 | **74.7±0.8** |

(2) Results of **Setting 3**.

The results under experimental Setting 3 are reported in Table IV, from which we can find that ELMs especially KELM outperform other methods. Due to that few labeled data from target domain are leveraged in model training with domain adaptation, so the recognition accuracies are much higher than that from Table III. The average differences between ELMs and SVMs are therefore reduced from 4% in **Setting 2** to 1.5% in **Setting 3**. For better visualization of the difference, we provide a Fig.4 which describes the recognition accuracies of all methods for each cross-domain task. We can see that KELM always shows the best performance.

TABLE IV
RECOGNITION ACCURACY OF EACH METHOD WITH **SETTING 3**, WHERE THE TRAINING DATA IS FROM BOTH SOURCE AND TARGET DOMAINS (A: AMAZON, C: CALTECH 256, W: WEBCAM, D: DSLR)

| Method | CNN_layer | A→D | C→D | W→D | A→C | W→C | D→C | D→A | W→A | C→A | C→W | D→W | A→W |
|---|---|---|---|---|---|---|---|---|---|---|---|---|---|
| NN | $f_6$ | 89.4±0.7 | 90.1±0.8 | 97.0±0.4 | 78.1±0.4 | 69.0±0.9 | 72.8±0.8 | 83.8±0.5 | 83.3±0.7 | 85.4±0.4 | 86.9±0.6 | 97.2±0.4 | 86.1±0.8 |
| | $f_7$ | 93.0±0.5 | 90.9±0.9 | 98.6±0.2 | 78.9±0.4 | 73.6±0.6 | 75.6±0.4 | 86.7±0.5 | 84.0±0.5 | 87.9±0.2 | 87.8±0.9 | 96.3±0.2 | 89.1±0.6 |
| SVM | $f_6$ | 94.5±0.4 | 92.9±0.8 | 99.1±0.2 | 84.0±0.3 | 81.7±0.5 | 83.0±0.3 | 90.5±0.2 | 90.1±0.2 | 90.0±0.2 | 91.5±0.6 | 97.9±0.3 | 90.4±0.8 |
| | $f_7$ | 94.0±0.6 | 92.7±0.8 | 98.9±0.2 | 83.4±0.4 | 81.2±0.4 | 82.7±0.4 | 90.9±0.3 | 90.6±0.2 | 90.3±0.2 | 90.6±0.8 | 98.0±0.2 | 91.1±0.8 |
| LSSVM | $f_6$ | 92.6±0.5 | 93.1±0.6 | 98.8±0.2 | 82.3±0.5 | 80.7±0.5 | 82.3±0.4 | 90.9±0.2 | 89.7±0.2 | 90.3±0.1 | 90.9±0.6 | 97.8±0.3 | 87.7±0.8 |
| | $f_7$ | 91.9±0.5 | 92.4±0.8 | 98.4±0.2 | 82.9±0.4 | 81.7±0.3 | 82.6±0.5 | 90.9±0.4 | 90.0±0.2 | 90.7±0.2 | 90.4±0.5 | 97.2±0.3 | 89.5±0.7 |
| ELM | $f_6$ | 94.6±0.5 | 93.7±0.6 | **99.2±0.2** | 83.4±0.3 | 81.2±0.3 | 83.5±0.3 | 91.1±0.2 | 90.3±0.2 | 90.5±0.1 | 91.6±0.7 | **98.3±0.2** | 90.5±0.6 |
| | $f_7$ | 94.9±0.4 | 93.0±0.6 | 99.0±0.2 | 84.1±0.2 | 82.2±0.4 | 84.1±0.2 | 91.7±0.2 | **90.8±0.2** | 90.9±0.1 | 91.5±0.7 | 97.9±0.2 | **91.7±0.7** |
| KELM | $f_6$ | **95.7±0.4** | 94.1±0.6 | **99.2±0.2** | 85.0±0.3 | 83.0±0.3 | 84.9±0.2 | 91.9±0.2 | 90.8±0.2 | 91.1±0.1 | 92.2±0.7 | 98.6±0.2 | 91.3±0.6 |
| | $f_7$ | 95.5±0.4 | 93.9±0.6 | 99.1±0.1 | **85.4±0.3** | **83.4±0.3** | **85.3±0.3** | **92.1±0.2** | **91.5±0.2** | **91.5±0.1** | 91.9±0.6 | 98.2±0.3 | **92.2±0.6** |

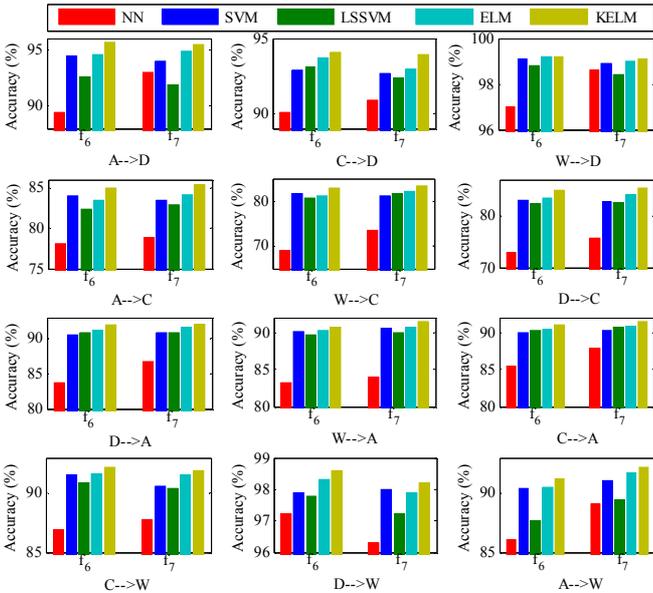

Fig. 4. Recognition accuracies of 12 cross-domain tasks by using NN, SVM, LSSVM, ELM and KELM on the deep convolutional activation features of $f_6$ and $f_7$ with experimental **Setting 3**

## V. CONCLUSION

In the report, we present a systematic comparison between SVMs and ELMs for object recognition with multiple domains based on the deep convolutional activation features trained by CNN on a subset of 1000-category images from ImageNet. We aim at exploring the most appropriate classifiers for high-level deep features in classification. In experiments, the deep features of 10-category object images of 4 domains from the 6-th layer and 7-th layer of CNN are used as the inputs of general classifiers including NN, SVM, LSSVM, ELM and KELM, respectively. The recognition accuracies for each method under three different experimental settings are reported. A number of experimental results clearly demonstrate that ELMs outperform SVM based classifiers in different settings. In particular, KELM shows state-of-the-art recognition performance among the presented 5 popular classifiers.

ACKNOWLEDGEMENT

This work was supported in part by the National Natural Science Foundation of China under Grant 61401048 and 61305144, in part by the Hong Kong Scholar Program under

Grant XJ2013044, and in part by the China Post-Doctoral Science Foundation under Grant 2014M550457.